# Keyword-Attentive Deep Semantic Matching

Changyu Miao[1] , Zhen Cao[1] and Yik-Cheung Tam[1]

[1] WeChat AI - Pattern Recognition Center, Tencent Inc.
{changyumiao, zhenzcao, wilsontam}@tencent.com

**Abstract.** Deep Semantic Matching is a crucial component in various natural language processing applications such as question and answering (QA), where an input query is compared to each candidate question in a QA corpus in terms of relevance. Measuring similarities between a query-question pair in an open domain scenario can be challenging due to diverse word tokens in the query-question pair. We propose a keyword-attentive approach to improve deep semantic matching. We first leverage domain tags from a large corpus to generate a domain-enhanced keyword dictionary. Built upon BERT, we stack a keyword-attentive transformer layer to highlight the importance of keywords in the query-question pair. During model training, we propose a new negative sampling approach based on keyword coverage between the input pair. We evaluate our approach on a Chinese QA corpus using various metrics, including precision of retrieval candidates and accuracy of semantic matching. Experiments show that our approach outperforms existing strong baselines. Our approach is general and can be applied to other text matching tasks with little adaptation.

**Keywords:** semantic matching, keyword dictionary, negative sampling.

## 1 Introduction

Open domain question answering (QA) is an important application ranging from web search to smart assistants [1,2,3,4]. One popular QA implementation is based on searching over a database that contains question and answer pairs. For fast search-ing speed, the database is often indexed by questions [5,6] so that similarity measures between an input query and a candidate question are defined based on TF-IDF scores or deep semantic matching that has gained a lot of attention in recent years [7,8,9,10]. Deep semantic matching is usually casted as a binary classification problem. A query pair is labeled as "positive" if the query pair is semantically similar or vice versa. However, we argue that the model may not learn well solely based on the binary class label. For instance, for a negative query pair "Which city is the capital of China?" and "Could you tell me the capital of America?", the model should figure out that their key differences are on "China" and "America". For open domain scenario, it is a big challenge to learn such differences/similarities due to combinatorial complexity of keyword/keyphrase pairs within the query pair. Especially when those keywords/keyphrases are "new" to models. Another issue in training a semantic matching model is the construction of positive and negative samples. Positive query pairs are usually obtained via retrieval of candidates followed by human validation. On the other hand, obtaining negative samples can be tricky as there are O(|Q|*|Q|) query pair combinations. Enumerating all negative samples are infeasible in terms of computational efficiency and the balance between positive and negative samples for effective model training.

In this paper, we address the above issues by designing keyword-attentive models and other supportive modules, including keyword extraction and negative sampling. We show that "keywords" are useful external signals to improve text matching, even if those keywords never appear in models' training data. Our main contributions are as follows:

1) Keyword-attentive BERT: We build our deep semantic model on top of the state-of-art BERT [11]. We introduce an extra keyword-attentive layer in parallel to the last layer of BERT. Our goal is to emphasize pairwise interaction between keywords and non-keywords in the attention mechanism [12]. By explicitly “telling” the model which word tokens are “important”, keyword-attentive BERT outperforms the original BERT in our experiments.

2) Robust model training via better negative sampling: To train a robust model, we propose a new negative sampling approach that employs a keyword overlapping score to select informative negative question pairs. In addition, we apply entity replacement trick to generate more variety of negative samples (e.g. replace the entity “China” by “America”). We will show that our model is more robust after training on the augmented data.

3) Keyword Extraction: We propose a simple and effective keyword extraction algorithm that leverages domain information. The extracted keywords can be used in three aspects: 1) To build a keywordattentive deep semantic matching model; 2) To improve retrieval quality in a QA search engine; 3) To improve negative sampling for training a better semantic matching model.

## 2 Related Work

There are many previous works related to query-question similarity tasks. Traditional methods such as retrieval incorporate mutual information [13] and term alignment scores [14] to measure pairwise interaction between the query-question pair. To better encode the pairs, many work focus on designing better network structures such as cross networks [15], convolution networks with kernel pooling [10] and adversarial networks [13,16], simpler network structure but more features [17], and attentive autoencoder to encode questions [18]. To deal with long text, text taxonomy is considered from word level to paragraph level [19]. To reduce the dependency of labeled training data, unsupervised or weakly supervised models were proposed based on a hash trick [20].

Injecting topic and keyword knowledge for better modeling is not uncommon. [21] trained an LDA model to predict topics as prior knowledge and designed a knowledge gate to leverage topics. [1] used semantic graphs to extract relations and perform reasoning. [22] proposed a keyword training structure using existing QA pairs and ap-plied keywords to product QA scenario. [23] directly used top-K words in a text sequence as keywords and formed a keyword mask in attention mechanism. Our approach differs [23] since we build upon the state-of-the art BERT with a keyword-attentive transformer layer.

For robust query-question model training, choosing the right samples is crucial. Random sampling is a common choice [6]. A better way is to measure negative question pairs using a distance score [24] or negative information [17]. Our approach introduces a notion of keyword overlapping score and keyword replacement to automatically generate informative negative samples.

# 3 Proposed Approach

## 3.1 Problem definition

Suppose we have a query q and the corresponding set of candidate questions . For each query pair (q, Qi), we calculate a similarity score sim(q, Qi) for candidate ranking. To calculate sim(q, Qi), we need to answer two questions: 1) How to easily and flexibly obtain good representation of the original query? 2) How to incorporate the query representation into the matching model?

To answer the first question, we propose a domain-based keyword extraction method to extract high-quality keywords in a query pair. To address the second question, we propose a keyword-attentive BERT to integrate keywords into the end-to-end model training.

## 3.2 Domain keyword extraction

Traditional retrieval methods such as TSUBAKI [25] or Elasticsearch [26] use OKAPI BM25 [27] or Lucene similarity [28] to measure the query-question distance. However, these methods may extract trivial "keywords" and result in low quality retrieval results. For example, as shown in Table 1, intuitively the important parts of the query are "中国 (China)" and "GDP". Our search engine tends to retrieve similar queries "Similar Q1" and "Similar Q2" that look very close to the original query but obviously ignore the mismatch on these important keywords.

**Table 1.** Examples of query retrieval results. (The keywords are in bold font)

| | |
|---|---|
| **Query** | 哪些因素会影响**中国**的 **GDP**<br>What factors will affect **China's GDP**? |
| **Similar Q1** | 哪些因素会影响**美国**的 **GDP**<br>What factors will affect **American's GDP**? |
| **Similar Q2** | **中国房价**的影响因素<br>Factors affecting **China's housing prices** |
| **True Q** | 说说**中国 GDP** 的影响因素<br>Talk about the factors affecting **China's GDP** |

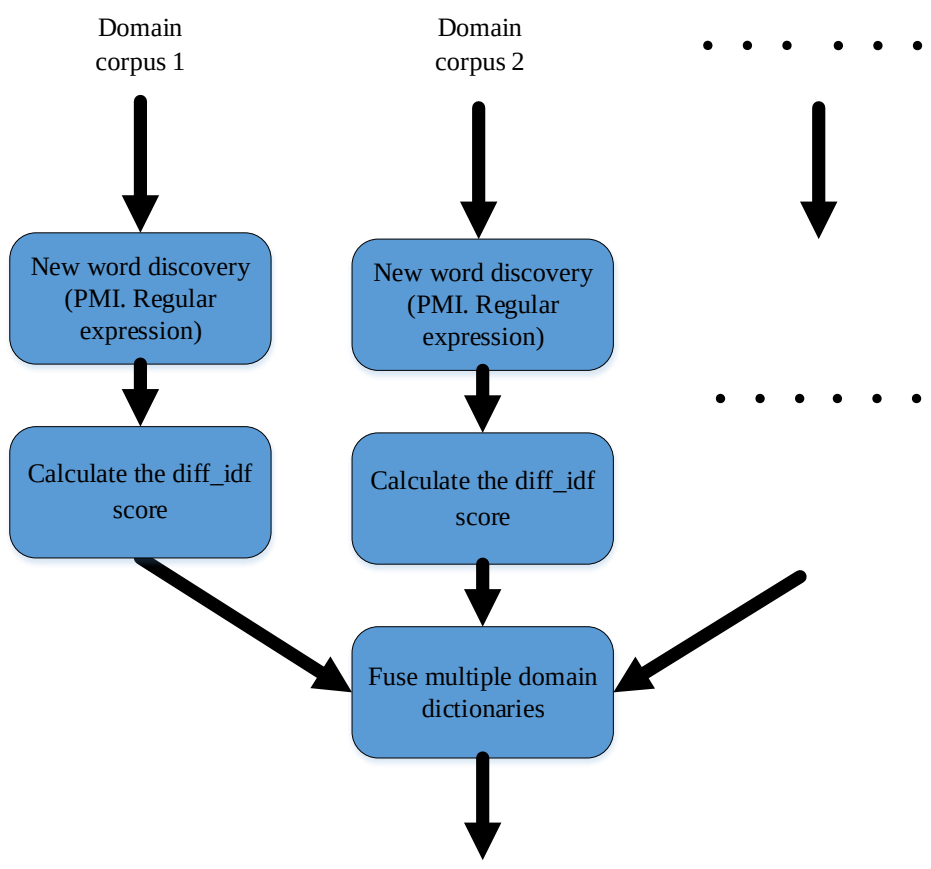

**Fig. 1.** Domain keyword extraction.

Actually, keywords of open-domain questions are highly related to questions' domains such as economy, politics, sports, etc. Based on such observations, we introduce a domain-based keyword extraction method as shown in Fig 1. to generate domain-related keywords. We've gathered large Chinese corpora containing tens of millions of articles belonging to certain domains and update them everyday to cover new keywords/keyphrases.

In Chinese NLP scenario, one Chinese "word" consists of several Chinese characters but has no white space as boundries, so word segmentations is a fundamental problem in Chinese NLP senario. Point-wise mutual information (*PMI*) is a common way to measure the degree of "stickiness" of neighboring tokens and find Chinese words:

$$PMI(w_1, w_2) = log\frac{p(w_1,w_2)}{p(w_1)p(w_2)}$$

In the second step, we measure the importance of a word by leveraging domain information. First, we compute *IDF* scores for each word. Then we introduce *diff-idf* score to measure the importance of a domain word as follows:

$$diff_idf(w|domain)$$
$$= idf(w|\hat{}domain) - idf(w|domain)$$
$$= log\left(\frac{N_{\hat{}domain}}{df(w|\hat{}domain) + \lambda}\right) - log(\frac{N_{domain}}{df(w|domain) + \lambda})$$

where ^domain indicates the anti-domain corpus. The reason why we use *df* instead of *tf* is that document-level frequency of a word in a domain corpus is more important than term-level frequency. The subtraction from the anti-domain *idf* promotes discrimination power of a word belonging to a target domain. Intuitively, if a word occurs evenly in articles across domains, its *diff-idf* score gets small due to the subtraction term. We use the *diff-idf* scores to sort the word candidates and remove noisy words that are below a threshold.

We repeat this procedure for each domain corpus, creating domain dictionaries. Finally, we merge domain dictionaries to form a final keyword dictionary and apply it to our search engine.

Compared to other unsupervised or supervised keywords extraction methods, our method has following advantages:

1) Could leverage domain information of corpora to extract domain keywords
2) No need to manually label keywords. No heavy model structures

### 3.3 Semantic matching

Our semantic matching approach is built upon BERT, by feeding a query pair into BERT [11] and creating a keyword-attentive layer as described below.

**Keyword attention mechanism**

After keyword extraction over a query pair, we inject keywords by stacking an additional keyword-attentive layer in parallel to the last layer of BERT as shown in Fig 2. Attention mechanism [29] is very important in semantic matching of query pairs. However, due to insufficient supervised signals, deep models may not accurately capture the key information in the query pair for effective similarity discrimination. Inspired by pair2vec [12], our proposed model pays more attention on the word-pair expression containing keywords. Specifically, suppose the input of the keyword-attentive transformer layer are $\{x_{CLS}^A, x_1^A, x_2^A \dots x_N^A, x_{SEP}^A, x_{CLS}^B, x_1^B, x_2^B \dots x_M^B, x_{SEP}^B\}$. As shown in Fig 3, each token in sentence A only attends to the keyword tokens in sentence B and vice versa. Sentence A and B are a negative (dissimilar) example. Sentence A & B look similar because they both contain "scan the QR code". But their meanings are different. A is about "joining a chat group" while B is "adding a new friend". Our keyword-attentive mechanism will enforce the model to learn their differences by focusing on the keyword difference between sentence A and B. This mechanism can be achieved easily by manipulating the self-attention masks in the transformer layer. Then average pooling is applied to the resulting hidden vectors of sentence A and B (with [CLS] and [SEP] excluded), forming two additional views of the sentence pair. To simulate the difference between these two representations, we introduce the keyword difference vector, *k-diff*, as follows:

$$k_diff = (h_{kw}(A) - h_{kw}(B)) \oplus (h_{kw}(B) - h_{kw}(A)) \qquad (3)$$

where $\oplus$ is a concatenation operator to concatenate the differences of average-pooled representations of A and B.

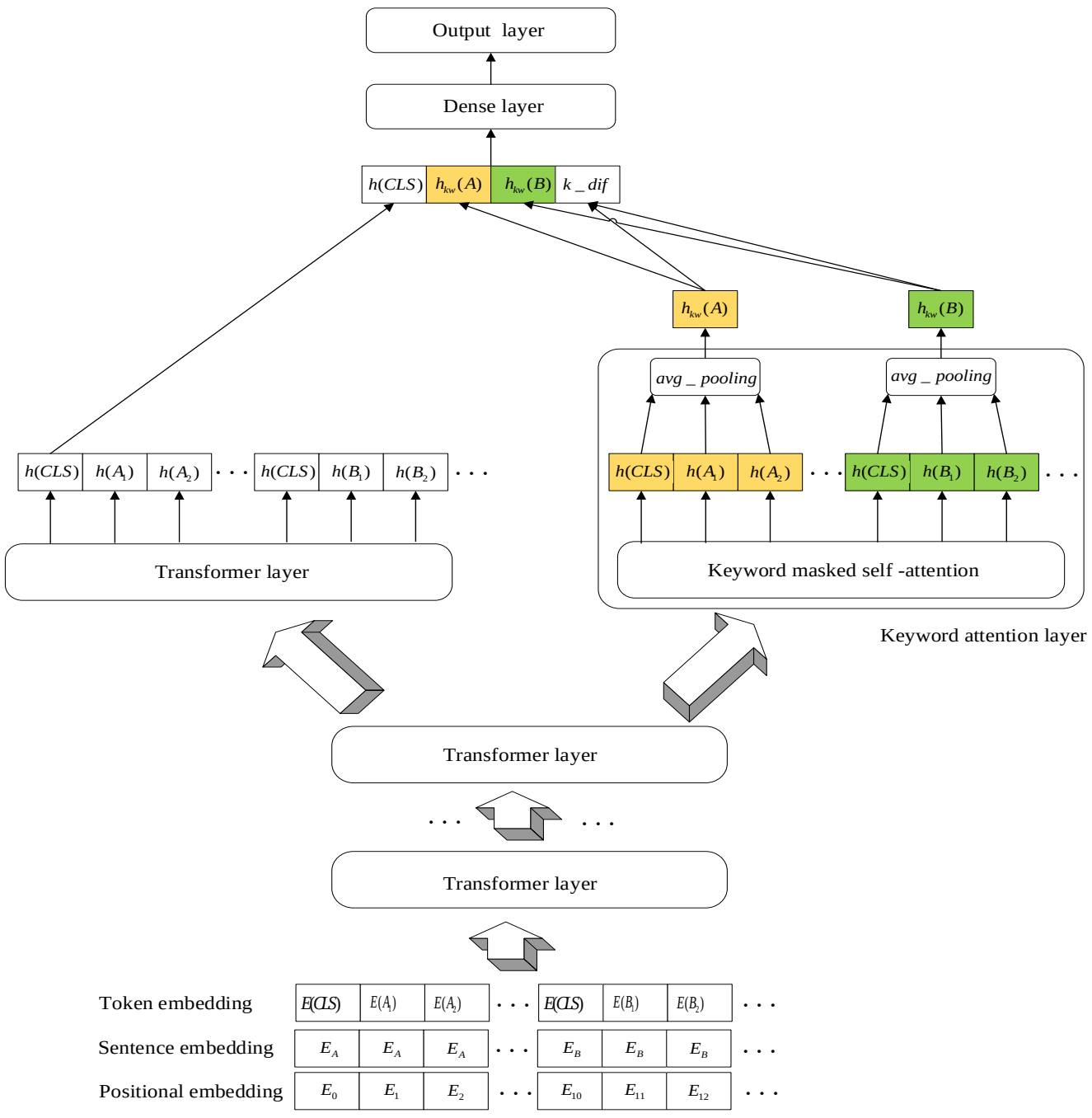

**Fig. 2.** Keyword attention mechanism. The green parts in sentence B are keywords and participate in attention from character tokens in sentence A.

With the keyword-attentive layer, the keyword information is injected closer to the output target rather than in the raw sequence input in BERT. Empirically, we found that the former approach is more effective and can keep the original BERT structure unmodified. Finally, we concatenate different views of query-pair representations for subsequent classification as follows:

$$h_{kw} = h(CLS) \oplus h_{kw}(A) \oplus h_{kw}(B) \oplus k_diff \qquad (4)$$

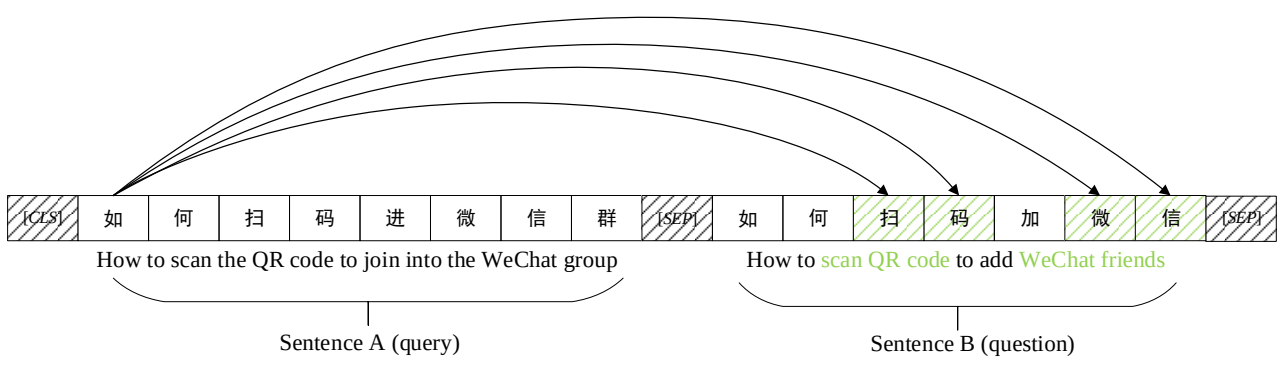

**Fig. 3.** Keyword attention mechanism. The green parts in sentence B are keywords and participate in attention from character tokens in sentence A.

## Negative sampling approach

As we cast semantic matching as binary classification, one issue is to create negative samples of query pairs. A simple solution is via random sampling to generate negative query pairs [6]. However, this solution is blind and tends to ignore informative samples. Even worse, it may just generate easy negative samples for classifier training. Therefore, our goal is to learn a robust model via better sampling. Motivated by support vectors, we want to generate confusing negative samples that have moderate similarity scores sim(q, Q). As a first step, we employ a search engine for retrieval using a keyword-augmented query. In particular, we augment the original input query with its keywords for candidate retrieval. Suppose the raw query is represented as a token sequence $\{x_1,x_2,\ldots x_n\}$ and the corresponding keywords are $\{k_1,k_2,\ldots k_m\}$. The search query is simply formulated as $\{x_1,x_2,\ldots x_n, k_1,k_2,\ldots k_m\}$.

Another question is how to pick the negative samples confidently from the search engine results without human supervision. One obvious metric is to check whether the retrieved candidate is confident according to the similarity score from a search engine. If the similarity score is below a threshold, then the retrieved candidate tends to be negative. In addition, we introduce the keyword overlapping ratio between an input query q and a candidate Q. Intuitively, a good negative sample pair should have a good balance between the non-overlapping and overlapping parts. Finally, we combine both criteria as a rule for negative-pair selection:

$$\begin{cases} 1, \text{ if } \dfrac{Sim(q,Q)}{Sim(q,q)} < \alpha \;\; AND \;\; \dfrac{|q \cup Q| - |q \cap Q|}{|q \cap Q|} > \beta \\ 0, \text{ otherwise} \end{cases} \tag{5}$$

where $\alpha$ and $\beta$ are hyper parameters for tuning. Here we find optimal $\alpha$ and $\beta$ to be 0.6 and 0.2 respectively by grid search. Sim(q, q) is considered as the maximum similarity for the query q itself returned by a search engine and is used for score normalization.

Another technique for generating negative samples is via random entity replacement. For instance, "What factors will affect China's GDP?" is rewritten automatically as "What factors will affect America's GDP?". This process generates query pairs that look very similar but certainly negative. Therefore, for each query in a database, we randomly replace one named entity inside the query according to a replacement ratio.

## 4 Experimental Setup

### 4.1 Data preparation

We constructed an open domain QA dataset via web crawling over Chinese QA community websites. We performed data preprocessing such as removing sensitive information, meaningless pairs using regular expression, and extremely similar questions that only differ by word order or punctation using a hash trick. The final dataset contained 100k QA pairs for indexing using Elasticsearch. In our experiments, we only indexed the questions in the search engine. For deep model training, we followed our negative sampling procedure described previously to generate the negative query pairs. To generate positive query pairs, we submitted a subset of training questions into Elasticsearch and picked the top-5 retrieved candidates of each question for human validation.

For accurate evaluation, we constructed a high quality Q-Q similarity test set via manual annotation. We random selected 1000 unseen questions from the preprocessed dataset and applied Elasticsearch on these questions yielding candidate questions. Then we asked human annotators to rewrite the retrieved candidates to generate positive and negative question pairs. To ensure unseen evaluation, we removed the rewrite questions that already existed in the database, yielding 818 positive and 946 negative question pairs for testing.

Details of the dataset to train and test the deep semantic model are shown in Table 2.

**Table 2.** Training and test sets for semantic similarity.

| Datasets | Number of question pairs (pos/neg) |
|---|---|
| Train (auto-gen) | 60000 (30000:30000) |
| Test (human) | 1764 (818/946) |
| Total | 61764(30818/30946) |

Currently, there're no public open-domain Chinese QA dataset for training and test, and also no Chinese corpora with domain tags for keywords extraction. We are willing to publish our well prepared dataset and source code on github for research purpose and for reproducibility.

### 4.2 Baseline models

For comparison, we chose a modified version of Fasttext, Fastpair, to mimic pair2vec, and a vanilla BERT as our baselines.

**Fastpair**

For classification, Fasttext [30] is a simple and efficient tool that can train on multi-core CPUs without expensive GPU machines and can yield competitive results compared to deep models. Fasttext employed a hashing trick to hash word N-grams of a query into embeddings. Average embedding is computed as the query representation, followed by linear projection and Softmax. We adapted Fasttext to become Fastpair for binary classification over a query pair. Inspired by pair2vec, the key of Fastpair is to introduce word-pair interaction features among the query pair, making the model suitable for query similarity task. Given a query q= $\{x_1^q, x_2^q, \cdots x_M^q\}$ and Q=$\{x_1^Q, x_2^Q, \cdots x_N^Q\}$, the input of Fastpair contains the following bag of word and cross word-pair features:

$$\{x_1^q, x_2^q, \cdots x_M^q\} \oplus \{x_1^Q, x_2^Q, \cdots x_M^Q\} \oplus \{x_i^q x_j^Q\},$$

$$\mathrm{i} \in [1, M], j \in [1, N]$$

Since Fastpair model is widely used in our productive environment, improving Fastpair does make sense from a productive point of view. For fair comparison with our proposed model, we integrated word-pair features that contain keywords. Suppose we extracted keywords over Q as Qkey $=\{x_1^{Q_{key}}, x_2^{Q_{key}}, \cdots, x_K^{Q_{key}}\}$. Then the additional pair features were added as $\{x_i^q x_j^{Q_{\mathrm{key}}}\}$, $\mathrm{i} \in [1, M], j \in [1, \mathrm{K}]$. Symmetrically, we swap q and Q above to create another set of pair feature. All features were then hashed into embedding buckets as performed in Fasttext.

**BERT**

We also used BERT as our deep model baseline. BERT is based on multiple layers of transformer to encode a text or a text pair. The success of BERT was due to the masked language model pre-training and next-sentence prediction over large text corpora, yielding good embedding representations. Through fine-tuning over a target labeled dataset, BERT has yielded state-of-the-art results over various tasks including sentence classification, sentence-pair classification, and question answering. We observed that BERT can achieve good accuracy even if the size of the target data set was not very large, thanks to its effective pre-training. Therefore, we applied BERT as our strong baseline. Here we choose a monolingual Chinese model published by Google as our pre-trained model.

### 4.3 Retrieval results

The baseline method uses the build-in BM25 method in the open-source elastic search. Our proposed method is by adding "keywords" into the index of elastic search. We retrieve the candidates using the test set from Table 2. Precision at different levels P@1, P@3, P@5, P@10 are used as our metrics. We first retrieved top-K candidates. If the reference question is contained in the top-K candidates, we assigned a score of 1. Table 3 showed that the keyword-enhanced searching yielded significantly better results at all precision levels. The extracted keywords effectively emphasized their importance in the query representation and guided the search engine for improved retrieval.

**Table 3.** Performance comparison of the retrieval system before and after adding the keyword information.

| Retrieval method | ***P@1*** | ***P@3*** | ***P@5*** | ***P@10*** |
|---|---|---|---|---|
| Baseline | 77.4% | 85.1% | 89.3% | 91.1% |
| Keyword-enhanced | 78.9% (+1.5) | 87.4% (+2.3) | 93.4% (+3.1) | 95.7% (+4.6) |

### 4.4 Semantic similarity results

Table 4 shows the overall test classification accuracy of different baselines on our semantic matching corpus. In general, Fastpair results were unsatisfactory probably due to the large number of parameters for the enumerated cross word pairs. BERT demonstrated a much stronger baseline. Our proposed keyword-attentive BERT showed significant gain over the BERT baseline. This agreed with our intuition that by pinpointing the keywords on the input query pair, our model was able to generate effective keyword-aware representations of the query pair in addition to the [CLS] representation.

**Table 4.** The performance comparison of retrieval system before and after adding the keyword information

| Method | ***Test Accuracy.*** |
|---|---|
| Fastpair | 72.3% |
| Fastpair (w/ keyword) | 78.5% |
| BERT | 93.9% |
| BERT (w/ keyword) | 95.1% |

Table 5 shows the overall test classification accuracy at various number of transformer layers. Our proposed model was significantly better than the strong BERT baseline at all layers. Even when 6 layers were used, our model achieved equal performance as the 12-layer BERT baseline at significantly reduced computation cost. This result was encouraging for launching a smaller and faster model in an online system.

**Table 5.** Test classification accuracies of BERT with various number of transformer layers.

| Number of layers | ***2*** | ***4*** | ***6*** | ***8*** | ***12*** |
|---|---|---|---|---|---|
| BERT | 87.7% | 90.4% | 92.0% | 93.5% | 93.9% |
| BERT (w/keyword) | 89.8% (+2.1) | 92.7% (+2.3) | 94.1% (+2.1) | 94.6% (+1.1) | 95.1% (+1.2) |

### 4.5 Negative sampling results

For negative sampling, we compared random sampling strategy with our proposed sampling strategy based on keyword overlapping ratio and entity replacement. Table 6 showed that random sampling would lead to much worse results compared to our proposed method. This was attributed to the fact that negative samples by random sampling were too easy to discriminate. This observation was also supported by the number of training epochs that led to convergence. The model training converged surprisingly fast even though only less than a half of the training pairs were observed. On the other hand, our proposed strategy produced more informative training samples and thus took more training epochs for convergence. We broke down the test set into positive and negative classes for analysis. Our model was slightly hurt on the positive test set but gained over 35% on the negative sets, showing that our proposed strategy was effective.

**Table 6.** Test classification accuracies with different negative sampling strategies.

| Method | Overall Acc | +ve class Acc | -ve class Acc | Training Epoch |
|---|---|---|---|---|
| Random | 77.5 % | 98.5 % | 56.5 % | 0.1~0.4 |
| w/keyword-overlap + entity replacement | 93.9 % (+17.8) | 96.2 % (-2.3) | 91.6 % (+35.1) | 2~3 |

### 4.6 Discussion

Compared to traditional BERT, our proposed keyword-attentive BERT has stronger ability to classify query-question pairs especially on very similar but negative query pairs. By introducing external domain-enhanced keyword dictionaries, our model knows which words are important in a query while traditional models have to learn those words themselves. The extra keyword-attentive layer helps better consider whether those important words will affect classification results. Table 7 shows some typical negative query-question pairs, which our keyword-attentive BERT correctly classified but traditional BERT failed.

**Table 7.** Negative query-question pairs correctly classified by keyword-attentive BERT but failed by the BERT baseline.

| Query | Question |
|---|---|
| 哪些因素会影响**中国**的**GDP**? | **中国房价**的影响因素? |
| What factors will affect **China**'s **GDP**? | Factors affecting **China's housing prices**? |
| **贸易战**对**中国房价**、**股价**有啥影响? | **贸易战**对**中美关系**有啥影响? |
| What effect does the **trade war** have on **China's housing prices** and **stock prices**? | What effect does the **trade war** have on **China's relation with America**? |
| **复联 3** 的故事**情节**是怎样的? | **复联 4** 的故事**情节**是怎样的? |
| What does the **story** in **Avengers 3** like? | What does the **story** in **Avengers 4** like? |
| 谁是 **NBA** 历史上**得分最多**的**球员?** | **NBA** 迄今为止**得分最多**的**球队**是哪支? |
| Which **basketball player** achieved the **highest score** in **NBA**’s history? | Which **team** in **NBA** achieved the **highest score** in ever since? |
| **美国乡村民谣**有哪些热门作品？ | 哪个城镇被称为**美国**的**音乐小镇**？ |
| Which songs are popular of **American’s country music**? | Which country in **America** is called “**a country with music**”? |

## 5 Conclusions

In this paper, we have proposed a keyword-attentive BERT for deep semantic matching. Our gains are attributed to effective injection of informative keywords into our model using keyword-attentive transformer layer to produce different keyword-sensitive representations of a query pair. For robust model training, effective negative sampling is very important. Empirically, we have shown that our proposed negative sampling approach based on keyword overlap and entity replacement outperforms simple random sampling. In the future, we will apply the proposed model in question answering with more deep features from a variety of semantic similarity models including question and answer similarity.